# Fall Detection from Indoor Videos using MediaPipe and Handcrafted Feature

Fatima Ahmed[1], Parag Biswas[2], Abdur Rashid[2], and Md. Khaliluzzaman[1]

[1]Dept. of Computer Science and Engineering, International Islamic University Chittagong (IIUC)
Chittagong-4318, Bangladesh
[2]Dept.of Master of Science in Engineering and Management, Westcliff University
California-90020, United Sates

fatimaahmediiuc@gmail.com, text2parag@gmail.com, rabdurrashid091@gmail.com, khalilcse021@gmail.com

*Abstract*—Falls are a common cause of fatal injuries and hospitalization. However, having fall detection on person, in particular for senior citizens can prove to be critical. Presently, there are handheld, ambient detector and vision-based detection techniques being utilized for fall detection. However, the approaches have issues with accuracy and cost. In this regard, in this research, an approach is proposed to detect falls in indoor environments utilizing the handcrafted features extracted from human body skeleton. The human body skeleton is formed using MediaPipe framework. Results on UR Fall detection show the superiority of our model, capable of detecting falls correctly in a wide number of settings involving people belonging to different ages and genders. This proposed model using MediaPipe for fall classification in daily activities achieves significant accuracy compare to the present existing approaches.

## I. INTRODUCTION

Falls are a leading cause of fatal injury, particularly in the elderly, and represent a significant barrier to independent living. Statistics about human fall detection reveal that the leading cause of fatal injury is a falling for an elderly person over 79 years old [1]. The most surprising issue is that the report made by WHO on the status of healthcare states that there will be twice as many gray heads (read senior citizens) over 60 years by 2050. Elderly trauma is associated with a high rate of complications, and falls have been identified as the single largest cause of hospitalization in this population [2]. On the other hand, China has an aging rate exceeding the rest of the world and will hit 35% worldwide by around 2025 compared with 2096 [3]. Hence, early identification, anticipation, and triggering an alert system in humans might be the most vital aspect of hospitalization.

There are three types of fall detection approaches overall: handheld device-based, ambient detector-based, and vision-based. For this method, the subjects are forced to wear the sensor device in order to receive signals data recorded (via a transceiver); it is still necessary for that data to be analyzed and characterized as a fall [4]. Next to the vision-based approach with purely one camera, this seems cumbersome and costly. For the computer vision part, we face several challenges as the body can deform during falls and artifacts remain present in the joint orientation absorption due to scale change. More-over, it was hard to detect a frame of all joint key points correctly. Also, when a person is really near the camera and its head is not in the frame during a fall situation, it is difficult to determine that it was a fall. For human fall detection, incorporate an image detection solution with MediaPipe for indoor environment videos. We have also evaluated our proposed model with one of the most common datasets, the UR fall dataset. The main contributions of this paper can be summarized as follows:

- Detecting as fall from human body skeleton for different ages and genders of people such as baby, young and old person in different indoor environment and complex situations with a handcrafted model using MediaPipe framework from the UR fall dataset.

- The model is evaluated with well-known UR Fall dataset to justify the effectiveness of the proposed model.

- Exploring this handcrafted model in indoor environment with significant accuracy.

The later part of this paper is well organized as follows— Section II represents the literature discussion on human fall events, Section III describes the methodology, Section IV dataset, Section V presents experimental details to evaluate the efficiency of the proposed model, Section VI discusses the efficiency of proposed method and VII describes the conclusion and potential future work.

## II. RELATED WORKS

Many researchers used many renowned models to detect and classify the fall detection from the different environmental conditions. Some of the models show the significant improvement. However, still there has enough room to improve the accuracy. In this paper, we basically focused on the hand-crafted feature-based fall detection.

Such as, first, Xueyi Wang et. al. [5] proposed a late fusion framework to aggregate the motion and spatial descriptors, as well as handcrafted features (e.g., HOG and LBP), optical flow calculated by an off-the-shelf pre-trained convolution neural network, and deep features. They design more than one experiment for determining the perfect camera location and process for both indoor and outdoor events. For assessing the generalizability of this approach, they perform leave-one-subject-out cross validation. After the tuning, the resulting trained model achieves 89.2% specificity, 93.6% sensitivity, and an overall accuracy of 91.8%.

Korumilli, Manasa, et al. [6] The important task of identifying unintentional falls that may result in serious injury or death is the topic of focus for. This is detected using key points that are associated with the coordinates on the human skeleton, and this uses Google's MediaPipe framework. A variety of classifiers, such as Random Forest (RF), Support Vector Machine (SVM) and a Deep Neural Network (DNN) model, are used to identify actions as either fall or non-fall. Experiments are also on the NTU-RGB+D dataset that performs real-time detection through webcam. The DNN model obtained the overall classification accuracy of 97.63%, whereas SVM and RF classifiers provided with an accuracy rate of 83.3% and 99.34% respectively.

Soman, Dawnee, et al. [7] proposed a vision-based fall detection system with MediaPipe technology, whereas the system is designed to inform the caregivers with a chatbot when their elderly people are falling. When positioned correctly so as to monitor the main field of activity, this improves the efficacy of monitoring assisted living situations with an elder. The MediaPipe model has been trained on falls as well, and a Random Forest method helps categorize different scenarios. It has an ID-based fall-type miner and is operating with a 90% true positive rate.

Bugarin, Charles Andrew Q. et al. [8] present a vision-based fall detection system that is executed on a smartphone app using deep learning, which is trained using multiple RGB camera configurations. Existing vision-based systems of fall detection suffer from the computational cost, the need for special cameras, and system compatibility with modern smartphones. This paper aims to solve these problems. The system utilizes MobileNetv2 CNN-based MediaPipe Pose and Random Forest Classifier for fall detection, with an on-premises auditory alarm, IoT notification, and real-time video feed in case of a fall. The model obtained 99.94% accuracy, clearly outperforming alternative vision-based approaches. All activities on the system performance—fall detection, indoor alarm, IoT message notification, and remote monitoring—were successfully validated with 100% accuracy in total.

Van-ha, Hoang et al. [9] have suggested the handcrafted feature-based approaches to improve deep learning algorithms. It reviews the latest skeleton-based fall detection methods using manual and deep learning in RGB videos. By giving brief introductions and insightful remarks, the paper offers a comprehensive performance comparison among the surveyed methods with state-of-the-art scores on multiple popular benchmark datasets. This paper discusses some potential future research directions and makes reference to the likely requirements and challenges with respect to practicing skeleton-based fall detection in RGB videos. This research paper aims to encourage the researchers working in this area to give more attention to issues related to the practical application of skeleton-based FDD systems—like user privacy and detection time cost—rather than focusing just on recognition accuracy. URFD had a result of 100% specificity, sensitivity, and total accuracy, while the UP-Fall resulted only at values of 88.7%, 92.94%, and 85.15%.

## III. RESEARCH METHODOLOGY

We have proposed a handcrafted method for human fall detection using MediaPipe pre-trained model. From UR Fall Video dataset, videos are passed through this model. This model detects the pose of human from the video frames and it detects the initial nose point. Based on a threshold fall is detected. After fall detection, it calculates the head angle, distance between nose and left ankle, percentage change of nose to left ankle with respect to initial nose point.

### A. MediaPipe Model

Human fall detection procedures are explained with Fig. 1.

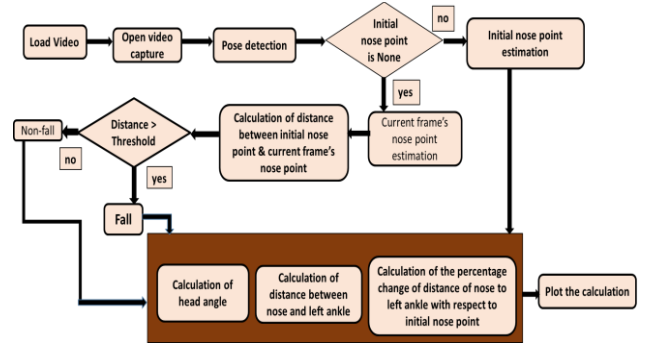

Fig. 1. Human fall detection procedures using MediaPipe pre-trained model

A video is a sequence frame. If the video ($V$) consists of $N$ frames, each frame $I_t$ ($t$ = frame number or time step) which is represented as:

$$V = \{I_1, I_2, I_3, \ldots, I_N\} \quad (1)$$

Here, a video capture object cv2.VideoCapure() from OpenCv library has used that can load and access these $I_t$ frames sequentially from the video file, each frame is a 2D or 3D matrix of pixel values. The frame $I_t$ can be represented as:

$$I_t(x, y) = [R(x, y), G(x, y), B(x, y)] \quad (2)$$

In equation (2), (x,y) represents the pixel location in each frame, R,G and B represent the red, green and blue channels respectively. These can be processed, analyzed and displayed for further uses.

A function cap.isOpened() checks if the file has been successfully opened or not. It mathematically can be represented as:

$$\text{isOpened} = \begin{cases} 1, & \text{if video is successfully opened}, t \leq N \\ 0, & \text{otherwise}, t > N \end{cases} \quad (3)$$

Another function cap.read() can read the next frame which was captured from the video and a boolean flag ret that indicates whether the frame was successfully read or not. It mathematically can be represented as:

$$(ret, I_t) = read(V, t), t \leq N \quad (4)$$

If ret = 1, the frame is successfully read or read fails ret = 0. If ret = 1, call the function PoseDetector(). The algorithm of pose detection has given below as:

| **Algorithm 1:** | | **Pose Detection** |
|---|---|---|
| **Input** | : | Image frame from video. |
| **Output** | : | The shape of the landmark (x, y) of that frame. |
| **Step 1** | : | Converts the image frames to RGB images. |
| **Step 2** | : | Estimates the pose of the RGB image and stores it into the results variable. |
| **Step 3** | : | Detects landmarks and extracts all the available key points. |
| **Step 4** | : | Append the shape of the landmark (x, y) to the points list and go to step 6. |
| **Step 5** | : | The function returns None and go to step 6. |
| **Step 6** | : | Stop. |

In pose detection algorithm, converts the image frame from the color space to the RGB color space which can be represented as:

$$I_{rgb}(x,y) = Convert(I_{bgr}(x,y)) \qquad (5)$$

From the RGB image, the pose is estimated and all the available key points are extracted using a CNN deep learning model. The model predicts 2D coordinates for each landmark which can be represented as:

$$L = f(I_{rgb}) \qquad (6)$$

In equation (6), f is the post estimation model. $L_i = (x_i, y_i)$ which represents the 2D coordinates of key body joints such as nose, neck, ankles etc. where i=1,2, 3, ..., N.

| **Algorithm 2:** | | **Fall Detection** | |
|---|---|---|---|
| **Input** | : | Image landmark points, initial head point, threshold. | |
| **Output** | : | Image detected fall or non-fall. | |
| **Step 1** | : | If | Points of landmark are not found and nose value > length(landmark) |
| | | | Return false. |
| | | Else End | Puts the nose value in head position. |
| **Step 2** | : | If | Initial head position and head position are found |
| | | | Calculates the distance between initial head position and head position. |
| | | Else if | Head distance > Threshold value, |
| | | | Returns the head distance. |
| | | Else | Returns false. |
| | | End | |
| **Step 3** | : | If | Returns head distance |
| | | | Draw a rectangle shape on the frame and (0, 0, 255) this value creates red colored the rectangle shape. |
| | | Else | Draw a rectangle shape on the frame and (0, 255, 0) this value creates the green colored rectangle shape. |
| **Step 4** | : | Stop. | |

In algorithm of fall detection, the potential falls are detected by calculating the displacement of nose between frames. The Euclidean distance between the current head position and initial head position is given below as:

$$d_{head} = \sqrt{(x_{initial} - x_{current})^2 + (y_{initial} - y_{current})^2} \qquad (7)$$

The fall is detected if the $d_{head}$ is greater than threshold = 95 pixels.

$$d_{head} > \text{threshold} \qquad (8)$$

| **Algorithm 3:** | | **Calculation of head angle, distance to ankle and percentage change** | |
|---|---|---|---|
| **Input** | : | Nose, neck and left ankle points. | |
| **Output** | : | Head angle, distance nose to ankle. | |
| **Step 1** | : | If | The key points of nose and neck are accessible |
| | | | Estimates the head angle ($\theta$) and assign the head angle in the variable head angle. |
| | | Else End | None |
| **Step 2** | : | If | Nose and left ankle key points are accessible |
| | | | Estimates the distance between nose and left ankle. |
| | | Else End | None |
| **Step 3** | : | If | Head angle and distance to ankles are evaluated |
| | | | Return head angle and distance to ankles values. |
| | | Else End | None |
| **Step 4** | : | Stop. | |

The head angle calculation algorithm measures the angle between the nose and the neck which is represented as:

$$\theta = \arctan\left(\frac{y_{nose} - y_{neck}}{x_{nose} - x_{neck}}\right) \quad (9)$$

The distance between the nose and left ankle is calculated by the Euclidian distance formula shown as:

$$d_{ankle} = \sqrt{(x_{nose} - x_{ankle})^2 + (y_{nose} - y_{ankle})^2} \quad (10)$$

When the initial distance is known, the percentage change in the distance between the nose and ankle can calculate which is represented as:

$$Percentage\ change = \frac{d_{ankle} - d_{initial}}{d_{initial}} \quad (11)$$

When no frame is returned, the video has ended and the loop breaks. This procedure of loop ensures that each frame has processed before ending of the video.

## IV. DATASET

Fall detection is now on treading as because falls are the leading cause of fatal and nonfatal injuries among the elderly. So now, it's a big issue to detect a person's fall accurately. Not only prevention but detection of falls as early as possible is very crucial for the health of the concerned person. Before choosing a dataset, we have to get the idea that how many ways a fall can be happened. A major injury depends on the action of fall and different fallen poster cause different kinds of fatal injuries. So, thinking about that we divide fall into many kinds [10].

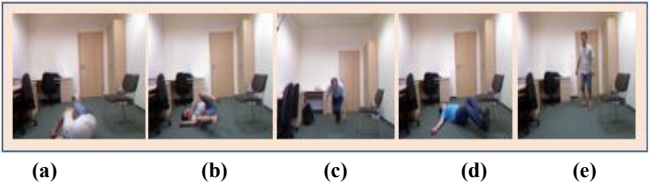

(a)    (b)    (c)    (d)    (e)

Fig. 3. Different types of falling from URFD dataset image: a) Left fall, b) Right fall, c) Forward fall, d) Backward fall and e) non-fall.

Here, the total number of videos in three illumination conditions such as normal, low, high that are processed through MediaPipe is 30*3=90 from the UR Fall dataset.

## V. EXPERIMENTAL RESULTS

### A. Results of handcrafted features

Detection of fall and non-fall, Calculation of head angle, distance to ankle and percentage from 3 illuminations condition in office environment:

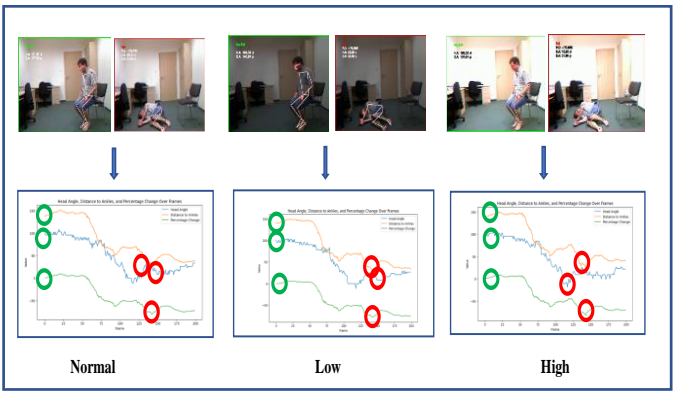

Normal     Low     High

Fig. 4. Sample video 01 is transformed into 3 illuminations normal, low and high. From every illumination fall and non-fall situations are classified with a complete skeleton of that person. Head angle (orange curve), Distance to ankle (blue curve), Percentage change (green curve) calculations are plotted into a graph for every frame in normal, low and high illuminations. From the video signals has generated according to calculated information. At the starting frame, the person is in non-fall situation which is showed (with green circle) in curve. When the person is in fall situation, we find the information of that situation in downward position of curve (red circle).

In Fig. 4, at first the normal, low and high illumination videos of the sample video 1 are uploaded. Then these are accessed through the equation no. (3) and when the frame is read through the equation (4), it converts the image to RGB using equation (5) and extracts thirty-three 2D coordinates of human body using equation (6). These points are represented as a complete skeleton.

Here, total 199 frames are available in sample video 1. We have picked 2 frames from every illumination condition where one is no-fall frame and another is fall frame detected using the equation (7), (8).

While the frame is detected as no-fall it makes a green border around the frame and the green colored text "No Fall" is showed on the frame. Otherwise, if the frame is detected as fall it makes a red border around the frame and the red colored text "Fall" is showed on the frame.

After showing the frame as fall and no-fall, the head angle, distance to ankle and percentage change are calculated using the equation (9), (10), (11) respectively and these 3 calculations of all frames are plotted and showed as orange, blue and green curves respectively.

In these curves, the green circles are showed for initial no-fall frame's position and the red circles are showed for next fall frame's position. While the curve is downward and a drastic change is seen, the fall frames are detected. Seeing the drastic changes of curves, we can also classify the fall situation.

TABLE I. CALCULATION OF HEAD ANGLE(DEGREE), DISTANCE TO ANKLE(PIXELS), PERCENTAGE CHANGE (%), THRESHOLD, CLASSIFICATION RESULT FROM SAMPLE VIDEO 01 IN THREE DIFFERENT ILLUMINATIONS LIKE NORMAL, LOW AND HIGH

| Illumination | Frame No. | Head Angle (H.A) degree | Distance to ankle (D.A) pixel | Percentage change (P.A) % | Threshold (pixel) | Detection (Fall=1, non-fall=0) |
|---|---|---|---|---|---|---|
| Normal | 0 | 101.31 | 137.28 | -- | 95 | 0 |
| Normal | 141 | 29.74 | 31.89 | -76.77 | 95 | 1 |
| Low | 0 | 101.31 | 140.87 | -- | 95 | 0 |
| Low | 141 | 23.20 | 33.30 | -76.36 | 95 | 1 |
| High | 0 | 102.53 | 137.91 | -- | 95 | 0 |
| High | 141 | 15.95 | 31.89 | -76.88 | 95 | 1 |

In this table, all these calculated information of 3 illumination conditions is collected from resultant frames and curves of sample video 1. From this table, we can see the head angle, distance to ankle and percentage change values of frames in normal, low and high illumination conditions are decreasing while falling.

## VI. DISCUSSION

Ten significant sequential frames from the total 199 frames for normal, low, and high illumination conditions of sample video 01 are represented below for better understanding.

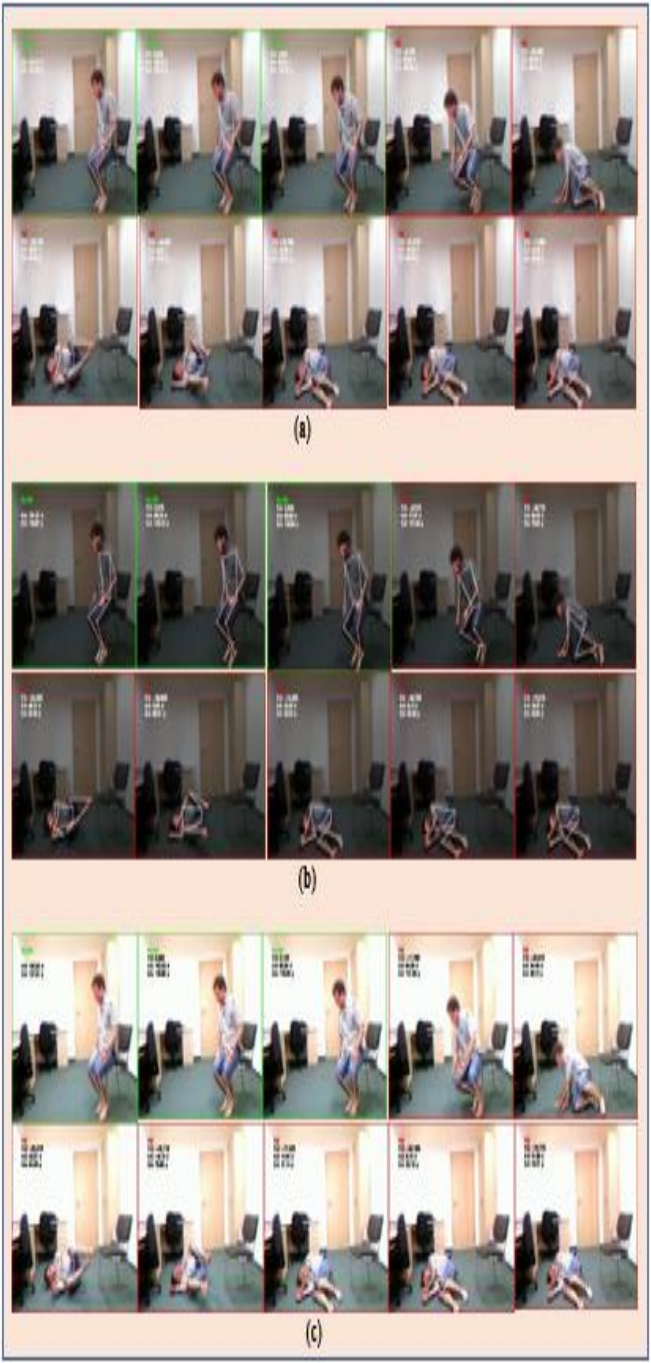

Fig. 5. Ten significant sequential frames (0, 20, 40, 60, 80, 100, 120, 140, 160, 199 no. frames) (a) from the normal illumination condition (b) from the low illumination condition (c) from the high illumination condition of sample video 01 from the UR Fall video dataset.

From Fig. 5, we can see the ten sequential frames (0, 20, 40, 60, 80, 100, 120, 140, 160, 199) from every illumination condition of sample video 01.

In these frames, the person is gradually falling down from a sitting position. When the head distance value of any frame is greater than 95 pixels, the person is detected as falling, with a red border around the frame and red text "Fall" on that frame. Otherwise, the person is detected as having no fall and a green border around the frame and green text "No Fall" on that frame.

After the observation of all these frames, we can understand MediaPipe can detect all fall and no fall frames with calculation of head angle, distance to ankle, and percentage change accurately. Only RGB data is used, not depth data. Our MediaPipe model is a pre-trained model. So, we have found all the 199 frames True Positive and no False Positive frame. In Fig. 5, ten significant sequential frames show how our pre-trained model detect all True Positive frames. Finally, we have got 100% accuracy.

Our proposed model shows a better accuracy than state of the art methods on UR Fall dataset. Youssfi Alaoui, et al. [11] in this paper, showed 96.55% accuracy on UR Fall Dataset using the Manifold of Positive Semidefinite Matrices and SVM. Jungpil Shin, et al. [12] in this paper, showed 97.53% accuracy on UR Fall Dataset using the Skeleton, Angle Feature and SVM. Finally, our proposed MediaPipe and Handcrafted method makes 100% accuracy on UR Fall dataset. Besides our proposed model has a low computational cost than other state of the art methods.

TABLE II. STATE OF THE ART COMPARISON

| Reference | Method | Dataset | Accuracy |
|---|---|---|---|
| [11] | Manifold of Positive Semidefinite Matrices +SVM | UR Fall | 96.55% |
| [12] | Skeleton +Angle Feature +SVM method | UR Fall | 97.53% |
| **Proposed** | **MediaPipe+ handcrafted method** | **UR Fall** | **100.00%** |

## VII. CONCLUSION

Fall detection is a general problem in computer vision, where the main objective is to find a person's body key point and detect the person is fall or not fall. We have proposed a method using MediaPipe framework that can detect falling of a single person in indoor environment, also it will estimate that person's body part length for the pose and the head angle, distance between head and ankle, percentage change the distance to ankle according to initial head for that person.

In this paper, we applied UR-Fall Dataset to find out a better accuracy. However, for successful deployment of such a design of a fall detection system in the real world, many aspects have to be considered in detail.

In MediaPipe method, we have some limitations. For example, from top view MediaPipe method is not able to detect fall. Our model is not able to detect multiple persons falling. In complex situation and different illumination condition, our defined threshold will be different which is another limitation.

In future, we want to overcome the limitations of our model. Besides we want to apply our method in both indoor and outdoor complex environmental situation and make a better accuracy and computational cost.

## REFERENCES

[1] M. Mubashir, L. Shao, and L. Seed, "A survey on fall detection: Principles and approaches," *Neurocomputing*, vol. 100, pp. 144–152, 2013.

[2] D. Wild, U. S. Nayak, and B. Isaacs, "How dangerous are falls in old people at home," *BMJ*, vol. 282, no. 6260, pp. 266–268, Jan. 1981.

[3] L. Yang, Y. Ren, H. Hu, and B. Tian, "New fast fall detection method based on spatio-temporal context tracking of head by using depth images," *Sensors*, vol. 15, pp. 23004–23019, 2015.

[4] X. L. Lau, C. T. Connie, M. K. O. Goh, and S. H. Lau, "Fall detection and motion analysis using visual approaches," *Int. J. Technol.*, vol. 13, no. 6, pp. 1173–1182, 2022.

[5] X. Wang, "Fall detection with a nonintrusive and first-person vision approach," *IEEE Journals & Magazines*, Sep. 19, 2023. [Online]. Available: https://ieeexplore.ieee.org. [Accessed: Feb. 10, 2024].

[6] M. Korumilli et al., "Human fall detection using skeleton features," in *Embedded Systems, Machine Learning and Signal Processing (PCEMS)*, 2023, pp. 1–6.

[7] D. Soman et al., "A novel fall detection system using Mediapipe," in *Proc. 2022 4th Int. Conf. Circuits, Control, Communication and Computing (I4C)*, 2022, pp. 123–127.

[8] C. A. Q. Bugarin et al., "Machine vision-based fall detection system using Mediapipe pose with IoT monitoring and alarm," in *Proc. 2022 IEEE 10th Region 10 Humanitarian Technology Conf. (R10-HTC)*, 2022, pp. 1–6.

[9] V.-H. Hoang et al., "Advances in skeleton-based fall detection in RGB videos: from handcrafted to deep learning approaches," *IEEE Access*, vol. 11, pp. 900–909, 2023.

[10] "UR Fall Detection Dataset," *fenix.ur.edu.pl*. [Online]. Available: https://fenix.ur.edu.pl. [Accessed: Dec. 5, 2024].

[11] A. Youssfi Alaoui, Y. Tabii, R. Oulad Haj Thami, M. Daoudi, S. Berretti, and P. Pala, "Fall detection of elderly people using the manifold of positive semidefinite matrices," *J. Imaging*, vol. 7, no. 7, p. 109, 2021.

[12] J. Shin, A. S. M. Miah, M. A. M. Hasan, Y. Okuyama, Y. Tomioka, "Fall detection using angle-based feature extraction from human skeleton and machine learning approach," in *Deep Learning and Visual Artificial Intelligence*, pp. 249–263, 2024.